\documentclass[sigconf]{acmart}

\usepackage{balance}
\usepackage{colortbl}

\def\BibTeX{{\rm B\kern-.05em{\sc i\kern-.025em b}\kern-.08emT\kern-.1667em\lower.7ex\hbox{E}\kern-.125emX}}
    
\setcopyright{acmcopyright}

\begin{document}

\copyrightyear{2019} 
\acmYear{2019} 
\acmConference[MM '19]{Proceedings of the 27th ACM International Conference on Multimedia}{October 21--25, 2019}{Nice, France}
\acmBooktitle{Proceedings of the 27th ACM International Conference on Multimedia (MM '19), October 21--25, 2019, Nice, France}
\acmPrice{15.00}
\acmDOI{10.1145/3343031.3350970}
\acmISBN{978-1-4503-6889-6/19/10}

\fancyhead{}

\title{Aesthetic Attributes Assessment of Images}


\author{Xin Jin}
\orcid{000-0003-3873-1653}
\affiliation{%
  \institution{Department of Cyber Security, Beijing Electronic Science and Technology Institute, Beijing 100070, P. R. China}
  \institution{CETC Big Data Research Institute Co.,Ltd., Guiyang 550018, Guizhou, P. R. China}
}

\author{Le Wu, Geng Zhao, Xiaodong Li, Xiaokun Zhang}
\affiliation{%
  \institution{Department of Cyber Security, Beijing Electronic Science and Technology Institute, Beijing 100070, P. R. China}
}

\author{Shiming Ge*}
\affiliation{\institution{Institute of Information Engineering, Chinese Academy of Sciences, Beijing 100093, P. R. China, \\*Corresponding author: }}
\email{geshiming@iie.ac.cn}

\author{Dongqing Zou}
\affiliation{\institution{SenseTime Research, Beijing 100084, P. R. China}}

\author{Bin Zhou}
\affiliation{
\institution{State Key Laboratory of Virtual Reality Technology \& Systems, Beihang University, Beijing 100191, P. R. China}
\institution{Peng Cheng Laboratory}
}

\author{Xinghui Zhou}
\affiliation{%
  \institution{Department of Cyber Security, Beijing Electronic Science and Technology Institute, Beijing 100070, P. R. China}
}

%
\renewcommand{\shortauthors}{X. Jin, et al.}

%
\begin{abstract}
Image aesthetic quality assessment has been a relatively hot topic during the last decade.  Most recently, comments type assessment (aesthetic captions) has been proposed to describe the general aesthetic impression of an image using text. In this paper, we propose Aesthetic Attributes Assessment of Images, which means the aesthetic attributes captioning. This is a new formula of image aesthetic assessment, which predicts aesthetic attributes captions together with the aesthetic score of each attribute. We introduce a new dataset named \emph{DPC-Captions} which contains comments of up to 5 aesthetic attributes of one image through knowledge transfer from a full-annotated small-scale dataset. Then, we propose Aesthetic Multi-Attribute Network (AMAN), which is trained on a mixture of fully-annotated small-scale PCCD dataset and weakly-annotated large-scale DPC-Captions dataset. Our AMAN makes full use of transfer learning and attention model in a single framework. The experimental results on our DPC-Captions and PCCD dataset reveal that our method can predict captions of 5 aesthetic attributes together with numerical score assessment of each attribute. We use the evaluation criteria used in image captions to prove that our specially designed AMAN model outperforms traditional CNN-LSTM model and modern SCA-CNN model of image captions.  
\end{abstract}

%
%


%

%

%
\begin{teaserfigure}
  \includegraphics[width=\textwidth]{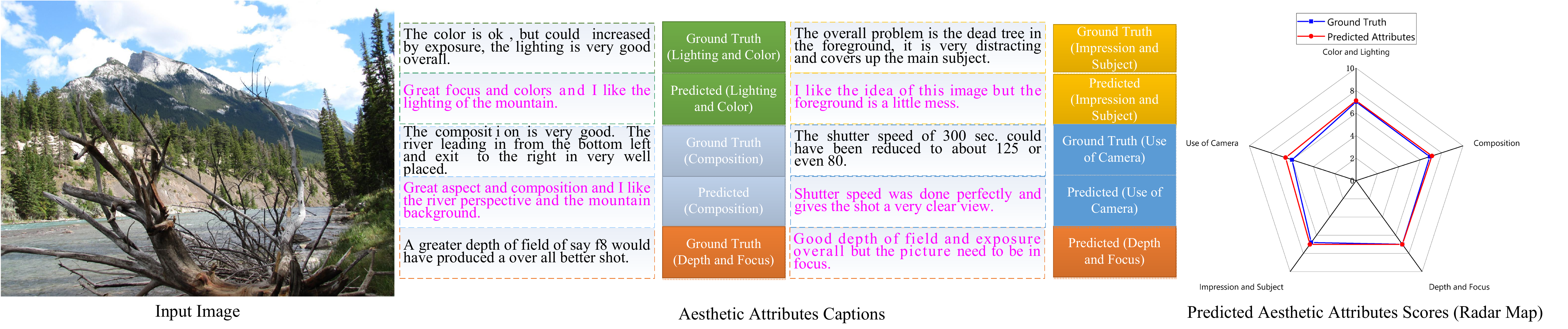}
  \caption{Aesthetic Attributes Assessment of Images. We predict caption and score of each aesthetic attribute of an image.}
  \label{fig:teaser}
\end{teaserfigure}

%
\maketitle
\section{Introduction}
Image Aesthetic Quality Assessment (IAQA) is to give an assessment of images on the aspect of aesthetics. In the last decades, IAQA has gained a great interest in the community of computer vision, computational aesthetics, psychology and neuroscience.

Most literatures of IAQA are to classify images into 2 categories: high aesthetic quality (professional) and low aesthetic quality (amateur). The second popular assessment task is to give a continuously numerical score of aesthetics. Another numerical assessment task is to predict a score distribution of human rating on the aesthetic aspect of an image  \cite{JinAAAI2018,TalebiTIP2018,CuiTMM2018}.

For a human artist, when shown a photo or a drawing, he/she will not just give a numerical score but always say a paragraph to describe many aesthetic attributes such as composition, lighting, color, focus of the image. Pioneer work of Chang et al. \cite{ChangICCV17} proposes aesthetic captioning of images. They build Photo Critique Captioning Dataset (PCCD) for the community. The PCCD contains 4,235 images and 29,645 comments. Each image is attached with comments and scores of 7 aesthetic attributes. However, they only output a sentence of assessment, which can not give a full review of aesthetic attributes. The value of PCCD is not fully explored. Besides, the size of PCCD is relatively small compared to AVA dataset \cite{MurrayCVPR2012}, which is commonly used in this field but do not contain ground truth of aesthetic captions and attributes.


In this work, we propose \emph{Aesthetic Attributes Assessment of Images}, as shown in Figure \ref{fig:teaser}. We predicts aesthetic attributes captions together with the aesthetic score of each attribute. We build a new dataset named \emph{DPC-Captions} from DPChallenge.com using an aesthetic knowledge transfer method. DPC-Captions contains comments of up to 5 aesthetic attributes of one image. There are 154,384 images and 2,427,483 comments. Then, we propose aesthetic multi-attribute network, which contains multi-attribute feature network, channel and spatial attention network, and language generation network. We train this model on both small-scale PCCD dataset (4,235 images and 29,645 comments) which contains attribute comments and scores and our large-scale DPC-Captions dataset with only contains attribute comments. We evaluate our method of captioning and scoring of attributes on DPC-Captions and PCCD using both image captioning criteria and mean square error of scoring. The contributions of our work includes:

\begin{itemize}
\item To the best of our knowledge, this is the first work which can produce both captions and scores for each aesthetic attribute of an image, including \emph{color and lighting}, \emph{composition}, \emph{depth and focus}, \emph{impression and subject}, \emph{use of camera}.

\item We introduce a novel large-scale image dataset named \emph{DPC-Captions} (154,384 images and 2,427,483 comments) for aesthetic assessment, which contains captions of up to 5 aesthetic attributes of images. The dataset building process relies on our proposed knowledge transfer method from a small-scale full annotated image dataset to a large-scale weakly annotated one.

\item We propose Aesthetic Multi-Attribute Network (AMAN), which uses a two-stage training processes on a small-scale full annotated dataset and a large-scale weakly annotated one. 
\end{itemize}

\section{Related Work}
Before deep learning era, many hand-crafted features \cite{JinECCV2010,ChenTIP2015} are designed for aesthetic image classification and scoring as surveyed by Deng et al. \cite{DengSPM2017}. Deep learning methods are proposed recently for aesthetic assessment \cite{JinAAAI2018,LuMM2014,DongNC2015,KaoSPIC2016,WangSP2016,MaiCVPR2016,KongECCV2016,JinWCSP2016,KaoTIP2017,MaCVPR2017}.
They outperform traditional methods. Lu et al. \cite{LuMM2014} present a two column CNNs which connects both local and global features for binary aesthetic classification. Mai et al. \cite{MaiCVPR2016} introduce ratio-preserving assessment of aesthetics by using SPP (Spatial Pyramid Pooling). Kong et al. \cite{KongECCV2016} propose the AADB (Aesthetics and Attributes Database) dataset which contains scores of 12 aesthetic attributes and use a rank-preserving loss for aesthetic scoring. Kao et al. \cite{KaoTIP2017} suggest a multi-task CNNs which can output results of both the binary aesthetic classification and multi-class semantic classification of an image. Jin et al. \cite{JinAAAI2018} present CJS-CNN (Cumulative Jensen-Shannon divergence) for aesthetic score distribution prediction. The aforementioned approaches only consider numerical assessment without taking the aesthetic assessment by languages into consideration.

\textbf{New tasks and datasets of IAQA}. Kong et al.  \cite{KongECCV2016} design a dataset named AADB (Aesthetics and Attributes Database) which contains 8 aesthetic attributes of each image. However, the label of each aesthetic attribute is only binary value (\emph{good} or \emph{bad}). Zhou et al.  \cite{AVAComments} design a dataset named AVA-Comments, which adds comments from DPChallenge.com to AVA dataset \cite{MurrayCVPR2012} which only contains aesthetic score distributions of images. Zhou et al. use the image and the attached comments to give a binary classification of aesthetics. Wang et al. \cite{AVAReviews} design a dataset named AVA-Reviews, which selects 40,000 images from AVA dataset and contains 240,000 reviews. Chang et al. \cite{ChangICCV17} design PCCD dataset, which contains 4,235 images and 29,645 comments. However, both \cite{AVAReviews} and \cite{ChangICCV17} can only give a single sentence as the comments of the aesthetics of an image . They do not describe the individual aesthetic attributes. 

\textbf{Image Captioning.} Most work of image captioning follow CNN-RNN framework and achieve great results \cite{KarpathyF17,DonahueHRVGSD17,MaoHTCY016}.
Most of recent literatures of image captioning \cite{Chen_2018_CVPR,Aneja_2018_CVPR,Anderson_2018_CVPR,Luo_2018_CVPR,Mathews_2018_CVPR}  introduce attention scheme. We follow this trend and add attention model in our network.

\begin{table*}[htb]
\centering
\caption{Aesthetic attribute keywords and frequency of PCCD dataset.}\label{tb:keywords}
\begin{tabular}{cc}
\hline
\textbf{Aesthetics Attributes} & \textbf{Top 5 Keywords (Frequency)} \\
\hline
Color Lighting & lighting (5829), color (5637), light (1708), sky (493), shadows (491)\\
\hline
Composition & composition (13749), left (2691), perspective (1787), shot (1715), lines (1369)\\
\hline
Depth of Field & depth (6087), field (5822), focus (1098), background (952), aperture (550)
\\
\hline
Focus & focus (7537), sharp (1308), eyes (402), see (345), camera (337)\\
\hline
General Impression & impression (4401), general (4357), good (1810), great (1338), nice (1040)\\
\hline
Subject of Photo & subject (6594), interesting (708), beautiful (386), light (209), capture (200)\\
\hline
Use of Camera & exposure (1619), speed (1488), shutter (1113), iso (1049), aperture (665)\\
\hline
\end{tabular}
\end{table*}

\begin{figure*}
\centering
\includegraphics[height=6.5cm]{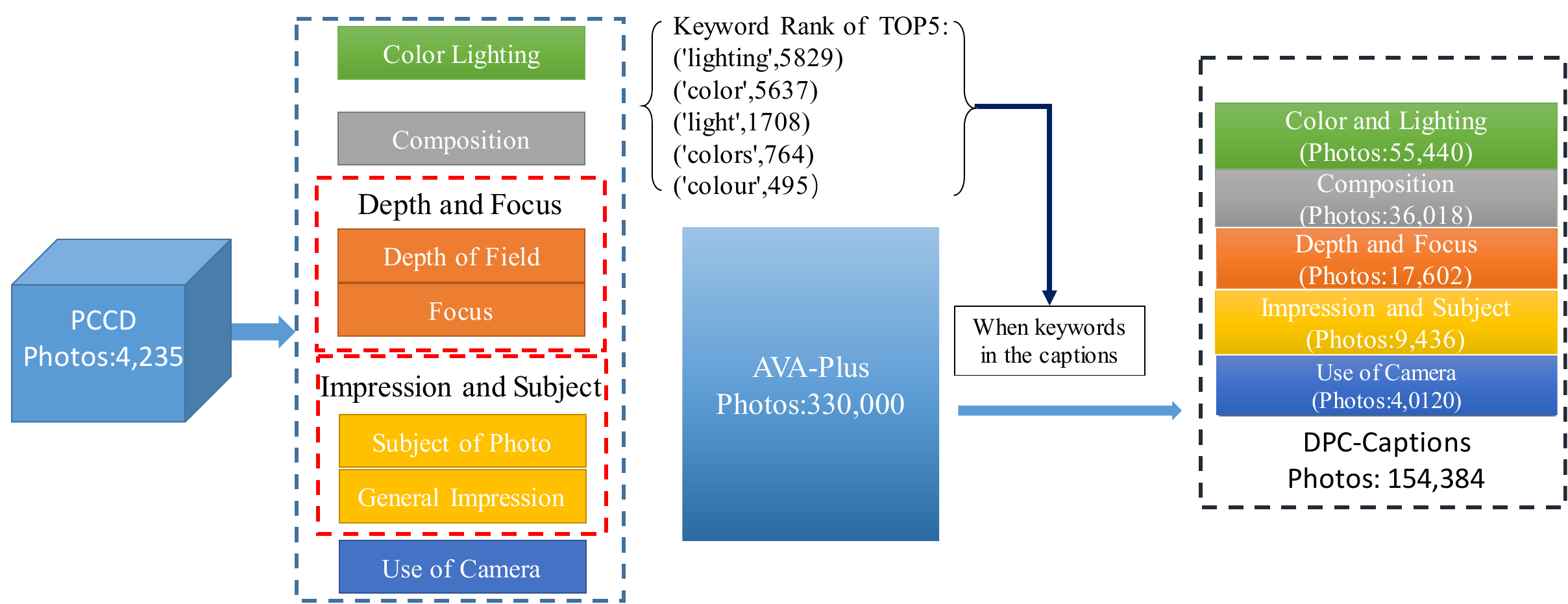}
\caption{The knowledge transfer method from PCCD to our DPC-Captions. The PCCD dataset includes 7 aesthetic attributes such as Color Lighting, Composition. The 5 keywords with the highest word frequency are selected from the comments of each aesthetic attribute. When a keyword appears in the comments of an image from DPC-Captions dataset, the image will be assigned to the corresponding aesthetic attribute. The repeated keywords make images be assigned into multiple attributes.}
\label{fig:dataset}
\end{figure*}

\begin{table*}
\caption{Comparison of different datasets. The average represents the number of  comments divided by the number of images.}
\label{tb:dataset} 
\centering
\begin{tabular}{lllll}
\hline\noalign{\smallskip}
Dataset & Number of Images & Number of Comments & Average & With Attributes\\
\noalign{\smallskip}\hline\noalign{\smallskip}
PCCD \cite{ChangICCV17} & 4,235 & 29,645 & 7 & \textbf{Yes}\\
AVA-Reviews \cite{AVAReviews} & 40,000 & 240,000 & 6 & No\\
AVA-Comments \cite{AVAComments}  & \textbf{255,530} & 1,535,937 & 6 & No\\
\textbf{DPC-Captions} & 154,384 & \textbf{2,427,483} & \textbf{15} & \textbf{Yes} \\
\noalign{\smallskip}\hline
\end{tabular}
\end{table*}


\section{DPC-Captions Dataset via Knowledge Transfer}
PCCD is a nearly fully annotated dataset, which contains comments and a score for each of the 7 aesthetic attributes (including overall impression, etc.). However, the scale of PCCD is quite small. While the AVA dataset contains 255,530 images with an assessment score distribution for each image. The images and score distributions of AVA dataset are crawled from the website of DPChallenge.com. Their exist comments from multiple reviewers attached for every image. However, the multiple comments are not arranged by aesthetic attributes.
We then crawl 330,000 image together with their comments from DPChallenge.com. We call this dataset AVA-Plus.

With the help of PCCD dataset \cite{ChangICCV17}, images of DPC-Captions are selected from the AVA-Plus. The aesthetic attributes of PCCD dataset include \emph{Color Lighting}, \emph{Composition}, \emph{Depth of Field}, \emph{Focus}, \emph{General Impression} and \emph{Use of Camera}. For each aesthetic attribute, keywords of top 5 frequency are selected from the captions. We omit the adverbs, prepositions and conjunctions. We combine words with similar meaning such as color and colour, color and colors. A statistic of the keywords frequency is shown in Table \ref{tb:keywords}.  

The selected aesthetic keywords such as \emph{composition}, \emph{color} and \emph{light} will become the core of DPC-Captions classification. For each image of AVA-plus (330,000), we label each comment which contains keywords in Table \ref{tb:keywords} using the corresponding attribute. We remove images whose comments do not contain keywords in Table \ref{tb:keywords}. Then, there remain 154,384 images of DPC-Captions. From the attribute view, we count images with each attribute. For the sake of balancing the number of images, we combine the \emph{Depth of Field} attribute with the \emph{Focus} attribute. We merge the \emph{Subject of Photo} attribute with the \emph{General Impression} attribute. Finally, we obtain 5 attributes of DPC-Captions, as shown in Table \ref{tb:keywords}. The number of images with \emph{Color and Lighting} is 55,440. While the number of images with \emph{impression and Subject} is 9,436, etc.

We compare our DPC-Captions with PCCD, AVA-Reviews, and AVA-Comments datasets in Table \ref{tb:dataset}. The AVA-Reviews and AVA-Comments do not contain aesthetic attributes. DPC-Captions has the most number of comments. Although we have less number of images than AVA-Comments, the average number of comments for each image in DPC-Captions is larger than that of AVA-Comments. For each attribute, we randomly select 2000 for validation and 2000 for testing. The remains are used for training.

\section{Multi-Attribute Aesthetic Caption}

\begin{figure*}
\centering
\includegraphics[width=0.85\textwidth]{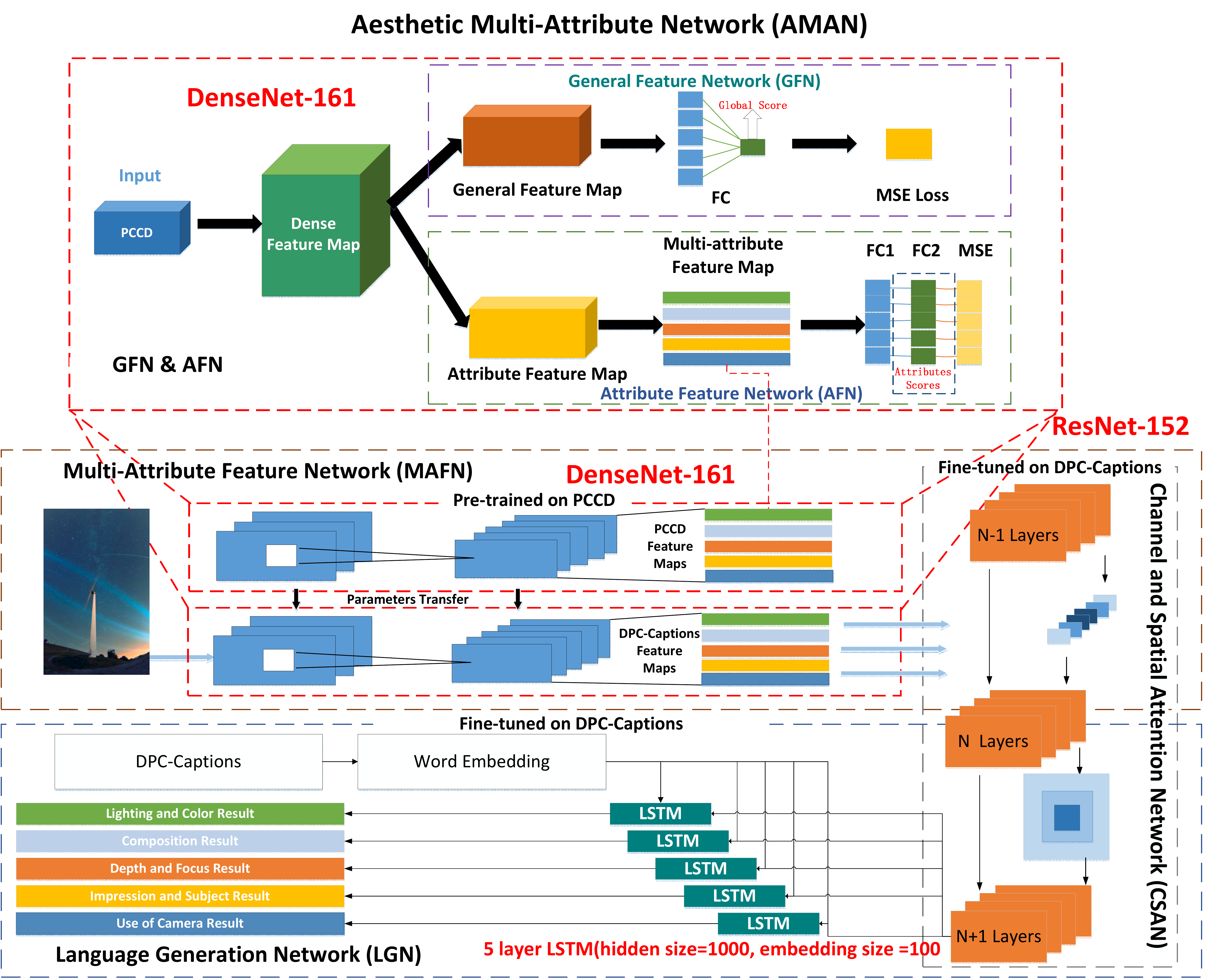}
\caption{Aesthetic Multi-Attribute Network (AMAN) contains Multi-Attribute Feature Network (MAFN), Channel and Spatial Attention Network (CSAN), and Language Generation Network (LGN). The core of MAFN contains GFN and AFN, which regress the global score and attribute scores of an image in PCCD using multi-task regression. They share the dense feature map and have separated global and attribute feature maps, respectively. Our AMAN is pre-trained on PCCD and finetuned on our DPC-Captions dataset. The CSAN dynamically adjusts the attentional weights of channel dimension and spatial dimension  \cite{SCACNNCVPR17} of the extracted features. The LGN generates the final comments by LSTM networks which are fed with ground truth attribute captions in DPC-Captions and attribute feature maps from CSAN.}
\label{fig:network}
\end{figure*}

The small-scale PCCD dataset contains both comments and scores of attributes. While our large-scale DPC-Captions dataset only contains attribute comments, which are selected from attached multi-user comments guided by the keywords of PCCD attribute comments. Besides, for images in DPC-Captions, maybe only part of the 5 attributes have attached comments.  We consider PCCD as a fully-annotated dataset, while DPC-Captions a weakly-annotated one. We propose to learn from a mixture of fully-annotated dataset and weakly-annotated dataset. 

The scores of attributes in PCCD are more powerful for extracting aesthetic features than just using comments. Thus, we leverage transfer learning both from scoring to comments of attributes, and from fully-annotated PCCD to weakly-annotated DPC-Captions. In addition, the attribute features are further enhanced by a channel and spatial attention model. At last, the enhanced attribute features are fed into LSTM to generate languages.

As shown in Figure \ref{fig:network}, the proposed Aesthetic Multi-Attribute Network (AMAN) is divided into three parts: Multi-Attribute Feature Network (MAFN), Channel and Spatial Attention Network (CSAN), and Language Generation Network (LGN). MAFN calculates the feature matrix of different attributes through the multi-task regression of 5 attribute scores. Due to the small scale of PCCD data and the full attribute labels, multi-attribute networks are be pre-trained on PCCD and fine-tuned on our DPC-Captions. CSAN dynamically adjusts the attentional weights of channel dimension and spatial dimension of the obtained features. Finally, LGN generates the captions by LSTM network which needs ground truth attribute captions in DPC-Captions and adjusted feature maps from CSAN.

\subsection{Multi-Attribute Feature Network (MAFN)}
Multi-task learning is a common method widely used in training deep convolutional networks. Due to the diversity of the attributes, multi-task learning can achieve multi-attribute assessment of aesthetics through parameter sharing. The aesthetic attributes assessment are relatively independent. However, the model training process is similar. In PCCD, in addition to scores for each attribute, there is a global score for each image. Thus, the loss of MAFN is divided into two parts. One is the loss of each attribute ($m$ attributes, in our paper $m=5$). The other is the global loss. $N$ represents the number of images in a batch. $\hat{y^{i}}$ represents the output of the last fully connected layer of the network. $y^{i}$ represents the true score. The equal sign in Eq. \ref{eq:lossAG} represents the same calculation method of the global loss and single attribute loss. There are totally 6 loss layers in this model.
\begin{equation} 
Loss^{Attribute}= Loss^{Gloal} =\frac{1}{2N}\sum_{i=1}^{N}\left \| \hat{y^{i}} - y^{i} \right \|_{2}^{2}
\label{eq:lossAG}
\end{equation}
\begin{equation}
Loss=\sum_{j=1}^{m}Loss^{Attribute}_j+Loss^{Gloal}
\label{eq:loss}
\end{equation}

As shown in the in the upper part of Figure \ref{fig:network}, the GFN and AFN use Desnet161 \cite{HuangLMW17} to extract the dense feature map. The parameters of all previous layers are shared. The output of GFN and AFN is divided into 6 parts: general feature and features of 5 aesthetic attributes. The GFN performs the full connected operation on the output of the global aesthetic score. For the final result, the calculation of the mean-square error (MSE) is performed and returned as a model loss parameter to the previous layers. The AFN performs the convolutional operation on the attribute feature map and get 5 different attribute feature maps. The same as the GFN, the final attribute scores are obtained through the full connection layer and the mean square error loss.

MAFN can extract different attribute feature maps of the image at the same time. Thus, the model is no longer limited to output comment of one sentence. The aesthetic characteristics of the image can be assessment from multiple attributes to better guide comprehensive assessment of images. The specific results obtained by the multi-task networks can also directly use the knowledge migrated to expand the attribute assessment of the DPC-Captions dataset, thus providing a broader aesthetic assessment ability.

\subsection{Channel and Spatial Attention Network (CSAN)}

The channel and spatial attention network \cite{SCACNNCVPR17} includes two modes. The first is the spatial attention after the attention of the channel. The second is the attention of the channel after the spatial attention. Through experiments, we use the first structure as our channel and spatial attention network part. Given the specific $N-1$ layer feature maps $M_{N-1}$, we obtain the channel attention weight $w_c$ according to the channel attention calculation $f_c$. We then linearly fuse the weight $w_c$ and $N-1$ layer feature maps to obtain new $N$ layer channel perceptual feature map s$M_{N}$. After that, the channel perceptual feature map $M_{N}$ is sent to the spatial perceptual attention module for operation $f_s$. The spatial attention weight $w_s$ is obtained. Finally, the channel perceptual feature map $M_{N}$ obtained in the previous step is spatially perceived which is the features of output from CNN. The process of merging can be expressed by the following formula.

\begin{equation}
f_{c}=tanh((w_{c}{\otimes}M_{N-1}+b_{c}){\oplus}w_{hc}h_{t-1})
\end{equation}

\begin{equation}
M_{N}=softmax(W_{N}f_{c} + b_{N})
\end{equation}

\begin{equation}
f_{s}=tanh((w_{s}M_{N-1}+b_{s}){\oplus}w_{hs}h_{t-1})
\end{equation}

\begin{equation}
M_{N+1}=softmax(W_{N}f_{s} + b_{N})
\end{equation}

In the above equations, $t$ represents the time state. $h$ represents the LSTM hidden state. $h_{t-1}$ records the hidden state of the last sequence. $\oplus$ represents the addition of the matrix and the vector. $\otimes$ represents the outer product of vectors. $b$ represents the offset.

\begin{figure*}
\centering
\includegraphics[height=15cm]{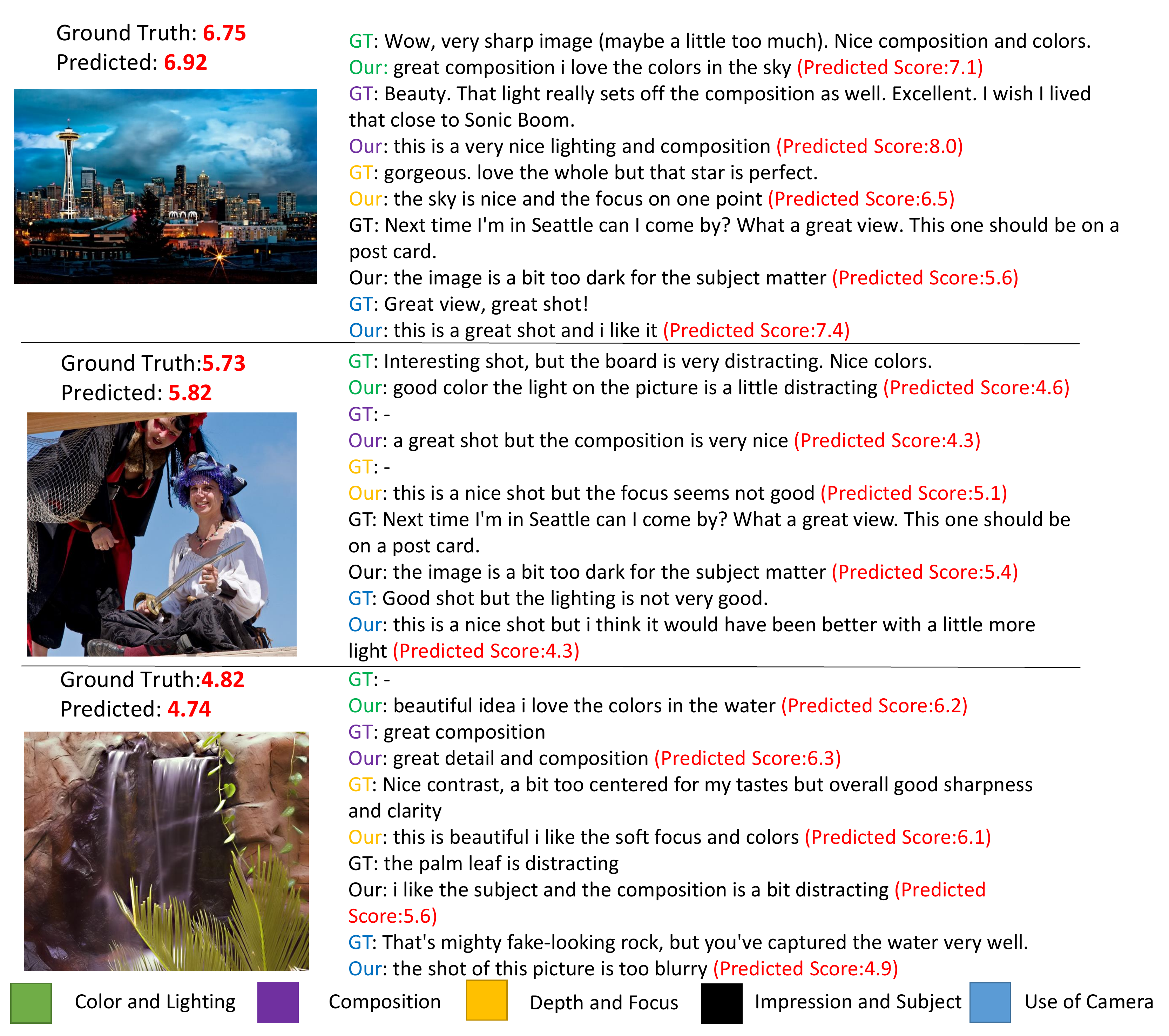}
\caption{The results of aesthetic multi-attribute network on DPC-Captions dataset. The predicted captions and score each attribute are shown. The Ground Truth score above each image is the global score from DPChallenge.com. The Predicted score above each image is the average score of the 5 predicted scores of attributes.}
\label{fig:ava}
\end{figure*}


\begin{table*}
\caption{Performance of the proposed models on the DPC-Captions dataset. We report RLEU-1,2,3,4, METEOR, ROUGE, and CIDEr. All values refer to percentage ($\%$). The method "without CSAN" is our AMAN without using CSAN part. The method "spatial first" is our AMAN using CSAN with spatial attention before channel attention.}
\label{tb:performance} 
\centering
\begin{tabular}{lllllllll}
\hline\noalign{\smallskip}
Method & BLEU-1 & BLEU-2 & BLEU-3 & BLEU-4 & METEOR & ROUGE & CIDEr \\
\noalign{\smallskip}\hline\noalign{\smallskip}
CNN-LSTM (Color and Lighting) & 46.3 & 23.2 & 13.6 & 6.9 & 12.5 & 27.0 & 6.1 \\
SCA-Model (Color and Lighting) & 46.5 & 23.3 & 13.9 & 7.1 & 12.8 & 27.4 & 6.2 \\
AMAN (Color and Lighting) without CSAN & 45.6 & 21.5 & 13.2 & 6.8 & 26.8 & 27.0 & 6.0\\
AMAN (Color and Lighting) spatial first & 44.1 & 19.2 & 10.1 & 6.7 & 25.3 & 27.0 & 6.0\\
AMAN (Color and Lighting) & \textbf{48.7} & \textbf{25.0} & \textbf{14.4} & \textbf{7.3} & \textbf{13.2} & \textbf{27.9} & \textbf{6.5} \\
\hline
CNN-LSTM (Composition) & 47.5 & 24.5 & 14.1 & 7.0 & 12.7 & 28.2 & 6.4 \\
SCA-Model (Composition) & 48.0 & 24.6 & 14.3 & 7.2 & 12.8 & 28.6 & 6.5 \\
AMAN (Composition) without CSAN & 47.4 & 24.5 & 13.5 & 6.1 & 10.6 & 28.1 & 6.4\\
AMAN (Composition) spatial first & 46.5 & 21.5 & 11.6 & 6.8 & 10.6 & 27.5 & 6.1\\
AMAN (Composition) & \textbf{49.2} & \textbf{24.9} & \textbf{14.6} & \textbf{7.5} & \textbf{13.6} & \textbf{28.9} & \textbf{6.8} \\
\hline
CNN-LSTM (Depth and Focus) & 46.2 & 23.1 & 13.2 & 6.4 & 12.2 & 26.8 & 6.0 \\
SCA-Model (Depth and Focus) & 46.3 & 23.3 & 13.4 & 6.5 & 12.3 & 26.9 & 6.0 \\
AMAN (Depth and Focus) without CSAN & 46.0 & 22.5 & 12.8 & 5.7 & 10.4 & 26.8 & 5.9\\
AMAN (Depth and Focus) spatial first & 36.4 & 14.9 & 4.2 & 2.3 & 6.6 & 19.4 & 3.3\\
AMAN (Depth and Focus) & \textbf{47.1} & \textbf{24.0} & \textbf{13.7} & \textbf{6.8} & \textbf{12.7} & \textbf{27.7} & \textbf{6.3} \\
\hline
CNN-LSTM (Impression and Subject) & 46.1 & 23.0 & 13.5 & 6.9 & 12.6 & 27.2 & 6.1 \\
SCA-Model (Impression and Subject) & 46.2 & 23.3 & 13.5 & 7.0 & 12.6 & 27.3 & 6.2 \\
AMAN (Impression and Subject) without CSAN & 45.7 & 22.9 & 12.8 & 6.9 & 12.6 & 27.2 & 6.0\\
AMAN (Impression and Subject) spatial first & 43.2 & 20.7 & 12.2 & 6.4 & 11.5 & 26.9 & 5.5\\
AMAN (Impression and Subject) & \textbf{46.8} & \textbf{23.6} & \textbf{13.9} & \textbf{7.4} & \textbf{13.0} & \textbf{27.6} & \textbf{6.7} \\
\hline
CNN-LSTM (Use of Camera) & 45.2 & 22.8 & 12.7 & 6.4 & 11.9 & 25.8 & 5.6 \\
SCA-Model (Use of Camera) & 45.3 & 22.9 & 12.9 & 6.4 & 12.1 & 25.9 & 5.7 \\
AMAN (Use of Camera) without CSAN & 44.9 & 22.8 & 12.7 & 6.4 & 12.0 & 25.8 & 5.7\\
AMAN (Use of Camera) spatial first & 34.4 & 13.0 & 4.5 & 2.4 & 6.7 & 17.0 & 2.0\\
AMAN (Use of Camera) & \textbf{46.2} & \textbf{23.4} & \textbf{13.2} & \textbf{6.7} & \textbf{12.8} & \textbf{27.2} & \textbf{6.5} \\
\noalign{\smallskip}\hline
\end{tabular}
\end{table*}

\begin{table*}
\caption{Performance of the proposed models on the PCCD test set. Results are compared through SPICE criterion \cite{SPICEECCV16}. The method "without CSAN" is our AMAN without using CSAN part. The method "spatial first" is our AMAN using CSAN with spatial attention before channel attention.}
\label{tb:spice} 
\centering
\begin{tabular}{llllll}
\hline\noalign{\smallskip}
Method & SPICE & Precision & Recall \\
\noalign{\smallskip}\hline\noalign{\smallskip}
CNN-LSTM-WD \cite{ChangICCV17} & 0.136 & 0.181 & 0.156 \\
AO Approach \cite{ChangICCV17} & 0.127 & 0.201 & 0.121 \\
AF Approach \cite{ChangICCV17} & 0.150 & 0.212 & 0.157 \\
\hline
CNN-LSTM (Color and Lighting) & 0.166 & 0.179 & 0.154 \\
SCA-Model (Color and Lighting) & 0.174 & 0.194 & 0.158 \\
AMAN (Color and Lighting) without CSAN & 0.165 & 0.177 & 0.154\\
AMAN (Color and Lighting) spatial first & 0.161 & 0.188 & 0.146\\
AMAN (Color and Lighting) & \textbf{0.196} & \textbf{0.231} & \textbf{0.170}  \\
\hline
CNN-LSTM (Composition) & 0.167 & 0.184 & 0.153 \\
SCA-Model (Composition) & 0.178 & 0.203 & 0.159 \\
AMAN (Composition) without CSAN & 0.165 & 0.183 & 0.154\\
AMAN (Composition) spatial first & 0.165 & 0.198 & 0.148\\
AMAN (Composition) & \textbf{0.197} & \textbf{0.228} & \textbf{0.174} \\
\hline
CNN-LSTM (Depth and Focus) & 0.163 & 0.174 & 0.153 \\
SCA-Model (Depth and Focus) & 0.167 & 0.182 & 0.154 \\
AMAN (Depth and Focus) without CSAN & 0.162 & 0.175 & 0.152\\
AMAN (Depth and Focus) spatial first & 0.095 & 0.107 & 0.082\\
AMAN (Depth and Focus) & \textbf{0.187} & \textbf{0.215} & \textbf{0.165} \\
\hline
CNN-LSTM (Impression and Subject) & 0.158 & 0.169 & 0.149 \\
SCA-Model (Impression and Subject) & 0.162 & 0.175 & 0.150 \\
AMAN (Impression and Subject) without CSAN & 0.157 & 0.170 & 0.151\\
AMAN (Impression and Subject) spatial first & 0.145 & 0.16 & 0.141\\
AMAN (Impression and Subject) & \textbf{0.181} & \textbf{0.213} & \textbf{0.158} \\
\hline
CNN-LSTM (Use of Camera) & 0.141 & 0.153 & 0.131 \\
SCA-Model (Use of Camera) & 0.154 & 0.167 & 0.143 \\
AMAN (Use of Camera) without CSAN & 0.142 & 0.155 & 0.131\\
AMAN (Use of Camera) spatial first & 0.041 & 0.098 & 0.084\\
AMAN (Use of Camera) & \textbf{0.176} & \textbf{0.209} & \textbf{0.156} \\
\noalign{\smallskip}\hline
\end{tabular}
\end{table*}


\begin{table}[htb]
\centering
\caption{The Mean Square Errors (MSE) of AFN and other methods on PCCD (described in the last paragraph of Section \ref{sec:comparisions}).}
\label{tb:afn}
\begin{tabular}{llll}
\hline
Attributes & \emph{regression} & \emph{multi-task} & AFN\\
\hline
Color and Lighting & 0.087 & 0.080 & \textbf{0.076}\\
\hline
Composition & 0.140 & 0.139 & \textbf{0.125}\\
\hline
Depth and Focus & 0.110 & 0.106 & \textbf{0.076}\\
\hline
Impression and Subject & 0.159 & 0.147 & \textbf{0.109}\\
\hline
Use of Camera & 0.223 & 0.193 & \textbf{0.128} \\
\hline
\end{tabular}
\end{table}

\subsection{Language Generation Network (LGN)}
Long Short Term Memory (LSTM) is a special type of RNN that learns long-term dependency information. In many problems, LSTM has achieved considerable success and has been widely used. By feeding information of  multiple attributes into the LSTM units, the prediction of the next word can be performed based on the image features and timing information. Specifically, if the two subtasks of the aesthetics assessment and the generated comment are unified, the training process can be described as this form: for a picture $I$ of the training set, the corresponding description is a sequence $S=\{S_1,S_2,...,S_N\}$ (where $S_i$ represents the sentence). For language generation model $\theta$ and attribute $a$, given the input picture $I$ , the probability of generating a sequence $S_i$ for each attribute as follow.

\begin{equation}
P_a(S|I)=\prod_{t=0}^NP_a(S_t|S_0,S_1,...,S_{t-1},I;\theta_a)
\end{equation}

The model utilizes the channel and spatial attention model to enhance the use of the effective area of the image. Thus, the features of the specific area of the image can be utilized more effectively in the decoder stage. The loss of the language generation network can be expressed by Eq. \ref{eq:lossa}, which controls the probability of the generated word vector by calculating the loss of each LSTM generating sequence.

\begin{equation}
Loss_{a}(I,S)=-\sum_{i=1}^{N}logP_{t}(S_{t})
\label{eq:lossa}
\end{equation}

The model uses the semantic information of the image to guide the generation of the word sequence in the decoder stage, avoiding the problem of using the image information only at the beginning of the decoder, which leads to the problem that the image information is gradually lost with time. In order to better obtain the high-level semantic information of the image, the model improves the original convolutional neural networks, including the method of multi-task learning, which can extract the high-level semantic information of the image and enhance the extraction of image features in the encoder stage.

\section{Experiments}

\subsection{Baseline}

\textbf{CNN-LSTM.} This model is based on Goolge's NIC model \cite{VinyalsTBE15}. The Resnet-152 \cite{HeCVPR2016} extracts features for different attributes and LSTM for encoding. The differences between this baseline and our method include: (1) no attention mechanism is introduced to enhance the feature extraction process; (2) the multi-tasking network is not used to extract features of different attributes. Instead, each attribute trains a network separately. It is not taking full advantage of the aesthetic features, we will carry out a simple knowledge transfer when extracting the characteristics of CNNs.

\textbf{SCA-Model.} This model is based on the SCA-CNN model \cite{SCACNNCVPR17}. The ResNet-152 extracts features for different attributes. LSTM performs spatial and channel attention enhancement features after extracting features. The differences between this baseline and ours include: (1) SCA-Model does not use multi-task networks to extract features of different attributes. Each attribute trains a network separately; (2) SCA-Model does not make full use of aesthetic features. A simple knowledge migration occurs when extracting features of CNNs.

\subsection{Implementation details}

Our experiments are based on Theano framework. The length of LSTM units is 1000. The features sent to the LSTM unit include 512-dimensional attribute features. The two stage training of AMAN is our contribution of using weakly supervised information. Except the two stage training, our AMAN, the baseline methods CNN-LSTM and SCA-Model use the same training parameters as follows: The word vector dimensionality is set to 50. The underlying learning rate is 0.01. The dimension of the force module and channel attention module is 512. The dropout is used in training to prevent overfitting. The network is optimized using a stochastic gradient descent optimization strategy. The batch size is set to 64 for DPC-Captions and 16 for PCCD. 

\subsection{Attribute Captioning Results}
We train and test our methods on the DPC-Captions and PCCD datasets. Some test results on the DPC-Captions dataset are shown in Figure \ref{fig:ava}. It is worth noting that the results we produce are not only rich in sentence structure, but also very accurate in grasping features. The relevance of comments and attributes are high. In terms of scoring, our average attribute score is very close to the ground truth score. Through the scores and comments, the evaluation of the image is vivid. \emph{The captioning and scoring results on PCCD dataset are shown in the supplementary materials due to the page limitation}. Our results can produce a variety of attribute results. The PCCD author's method \cite{ChangICCV17} can only produce one sentence. In addition, our results tend to be objectively evaluated, and the PCCD author's approach favors subjective evaluation.

\subsection{Comparisons}
\label{sec:comparisions}

\textbf{Comparison with baseline.} The evaluation criteria to compare the performance of our model and the baseline models include RLEU-1,2,3,4, METEOR, ROUGE, and CIDEr, which are commonly used in nature language processing community. The comparison results shown in Table \ref{tb:performance} reveal that our model outperforms the baseline models in all criteria. The Use of Camera and Impression and Subject attributes are not as good as the first three attributes. The number of comments of these two attributes is relatively small.

\noindent\textbf{Comparison with other methods.} We use SPICE \cite{SPICEECCV16} to compare the performance between the methods \cite{ChangICCV17} and our model. SPICE is a criteria for the automatic evaluation of generated image captions. It resolves the similarity between the result and the generated captions by parsing the sentence into a graph. The calculation formula is as follows.

\begin{footnotesize}
\begin{equation}
SPICE=F_{1}Score=\frac{2*Precision*Recall}{precision+Recall}
\end{equation}
\end{footnotesize}

As shown in Table \ref{tb:spice}, our model is superior to the method proposed by the PCCD \cite{ChangICCV17} in various attributes. The PCCD method \cite{ChangICCV17} uses the attribute fusion training method, which combines the three attributes of Composition, Color and Lighting, Subject of Photo. However, by contrast it can be found that the comments we generate in these three properties have better comments than the previous ones.

\textbf{Attributes scoring.} To compare the performance of aesthetic attributes scoring, we compare AFN (Attribute Feature Network) with other methods on PCCD, as shown in Table \ref{tb:afn}. The \emph{regression} method uses DenseNet161 \cite{HuangLMW17} to perform simple regression on scores without adding a multi-attribute structure. The \emph{multi-task} method uses a multi-attribute combination method but does not use a branch structure of global scores. Obviously, our approach has a big advantage in predicting the scores of individual attributes.

\section{Conclusion and Discussion}
In this paper, we propose a new task of IAQA: aesthetic attributes assessment. A new dataset called DPC-Captions is built through knowledge transfer for this task. We propose a novel network AMAN for two-stage learning processes on both full annotated small-scale dataset and weakly annotated large-scale dataset. Our AMAN can generate captions and scores of individual aesthetic attributes. 

In the future, we will explore to caption from sentences to paragraphs. The knowledge transfer methods can be used to build larger dataset for weakly supervised learning. The relations among  attributes can not only be used for scoring learning but also for caption learning. Reinforcement learning can also be leveraged for captions generation.

\section{Acknowledgments}

We thank all the ACs and reviewers. This work is partially supported by the National Natural Science Foundation of China (Grant Nos. 61772047, 61772513), the Open Project Program of State Key Laboratory of Virtual Reality Technology and Systems, Beihang University (No. VRLAB2019C03), the Open Funds of CETC Big Data Research Institute Co.,Ltd., (Grant No. W-2018022), and the Fundamental Research Funds for the Central Universities (Grant Nos.328201907).

{\small
\bibliographystyle{ACM-Reference-Format}
\bibliography{acmart}
}

\end{document}